\documentclass{article}
\usepackage{ijcai25}
\usepackage{amsmath,amsfonts}
\usepackage{subfigure}
\usepackage{times}
\usepackage{soul}
\usepackage{url}
\usepackage[hidelinks]{hyperref}
\usepackage[utf8]{inputenc}
\usepackage[small]{caption}
\usepackage{graphicx}
\usepackage{amsmath}
\usepackage{amsthm}
\usepackage{booktabs}
\usepackage{algorithm}
\usepackage{algorithmic}
\usepackage[switch]{lineno}
\usepackage{multirow}

\usepackage{bbm}
\usepackage{multirow, multicol}
\usepackage{arydshln}
\usepackage{xcolor}
\definecolor{table_row}{rgb}{0.90,0.90,0.91}
\usepackage{colortbl} 
\usepackage{tabularx}
\usepackage{subcaption}
\usepackage{bbding}
\usepackage{amssymb}

\newcommand{\softmax}{\operatorname{softmax}}
\newcommand{\argmax}{\operatorname{argmax}}

\newcommand{\logg}{\operatorname{log}}
\newcommand{\bigp}{\mathbf{P}}

\newcommand{\smallp}{\mathbf{p}}

\title{Concentrate on Weakness: Mining Hard Prototypes for Few-Shot Medical Image Segmentation}

\author{
Jianchao Jiang\And
Haofeng Zhang\thanks{Corresponding author.}
\affiliations
School of Computer Science and Engineering, Nanjing University of Science and Technology, China\\
\emails
\{jiangjianchao, zhanghf\}@njust.edu.cn
}

\begin{document}

\maketitle

\begin{abstract}
Few-Shot Medical Image Segmentation (\textbf{FSMIS}) has been widely used to train a model that can perform segmentation from only a few annotated images. However, most existing prototype-based FSMIS methods generate multiple prototypes from the support image solely by random sampling or local averaging, which can cause particularly severe boundary blurring due to the tendency for normal features accounting for the majority of features of a specific category. Consequently, we propose to focus more attention to those weaker features that are crucial for clear segmentation boundary. Specifically, we design a Support Self-Prediction (\textbf{SSP}) module to identify such weak features by comparing true support mask with one predicted by global support prototype. Then, a Hard Prototypes Generation (\textbf{HPG}) module is employed to generate multiple hard prototypes based on these weak features. Subsequently, a Multiple Similarity Maps Fusion (\textbf{MSMF}) module is devised to generate final segmenting mask in a dual-path fashion to mitigate the imbalance between foreground and background in medical images. Furthermore, we introduce a boundary loss to further constraint the edge of segmentation. Extensive experiments on three publicly available medical image datasets demonstrate that our method achieves state-of-the-art performance. Code is available at \url{https://github.com/jcjiang99/CoW}.
\end{abstract}
\section{Introduction} \label{sec:introduction}
	\begin{figure}[!t]
		\centering
		\includegraphics[width=0.48\textwidth]{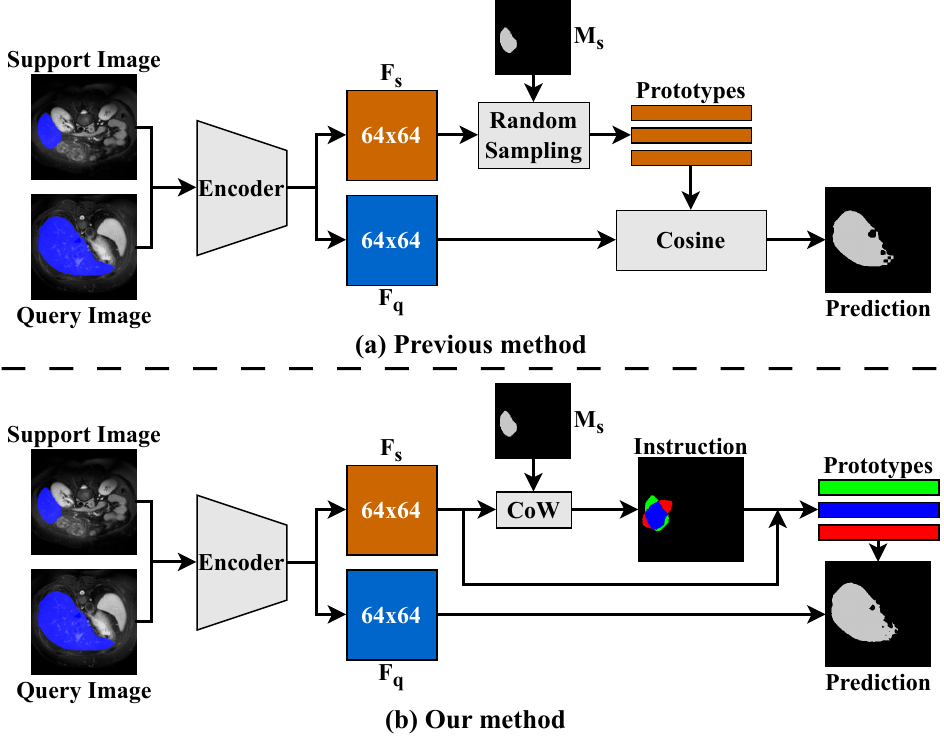}
		\caption{Comparison between previous methods and ours. (a) The previous methods generate multiple prototypes only by random sampling or local averaging, giving insufficient attention to weak boundary details. (b) Our proposed CoW instead concentrates more on these weak features of a specific category, thus acquiring a more comprehensive representation of the category distribution and achieving clearer segmentation boundary.}
		\label{fig_motivation}
	\end{figure}

Automatic segmentation of organs, tissues and lesions appearing in medical images is broadly applied in clinical research, including disease diagnosis \cite{diag1}, treatment planning \cite{treat1,treat2,treat3} and surgical guidance \cite{surgi1,surgi2}. Recently, fully supervised deep learning segmentation frameworks \cite{super1,super2,super3} exhibit superior performance, but they heavily depend on large amounts of densely annotated training data, which is impractical for medical scenarios for the reason that labeling medical images requires considerable clinical expertise, cost and time \cite{label1}. 

To solve this challenge, Few-Shot Learning (FSL) \cite{fsl1,fsl2,fsl3,fsl4} has been proposed as a potential learning paradigm to train models that can efficiently adapt to novel classes when given extremely scarce annotated data \cite{fss1}, based on which corresponding segmentation method is called Few-Shot Medical Image Segmentation (FSMIS). In particular, FSMIS typically trains models with base classes and segment unseen classes in the inference stage. Consequently, models trained in this manner are capable of segmenting novel organ or lesion leveraging solely a few annotated data, evading the need for retraining, which holds significant importance for medical image segmentation. 

As one of dense image prediction tasks, FSMIS has an indispensable demand for features with strong category information and precise spatial boundary details. In order to leverage the annotated samples, the majority of current FSMIS models distill prior knowledge based on prototype network \cite{ALPNet,QNet,DCLAAS}, among which multiple prototypes-based network is the predominant method. However, these methods generate multiple prototypes from support images either by a random selecting strategy or simple local averaging, which incline to normal features in the central area of a specific organ or involve background information in foreground prototypes as illustrated in Fig. \ref{fig_motivation}(a). Furthermore, those weaker features mainly distributed along the boundary of the organ are neglected to a considerable extent, which can lead to a poorly segmentation boundary. As a result, some foreground and background points will be misclassified due to the excessive focus on normal features.

In this paper, we propose a Concentrate on Weakness (CoW) approach that learns to mine hard prototypes for each category to represent its feature distribution more comprehensively as illustrated in Fig. \ref{fig_motivation}(b), to tackle the aforementioned problem. Specifically, we devise a Support Self-Prediction (SSP) module to provide knowledge for weak features carrying critical boundary information, which are subsequently utilized by the Hard Prototypes Generation (HPG) module to generate hard prototypes to improve segmentation accuracy and robustness. Right after is the Multiple Similarity Maps Fusion (MSMF) module that fuses multiple similarity maps to incorporate comprehensive information included in them to get the final predicted mask. To acquire a more precise segmenting boundary, we also devise a boundary loss, which effectively suppress over-segmentation whether for the foreground or the background. In conclusion, our main contributions are summarized as follows:

\begin{itemize}
    \item We propose a novel method to identify weak features by comparing predicted support mask and its true label. Besides, multiple hard prototypes are generated by further processing those weak features, ensuring more comprehensive representation of overall distribution. 
    \item To make full use of the information provided by generated multiple prototypes, we design an MSMF module to fuse multiple similarity maps predicted by these prototypes to get the final mask.
    \item Extensive experiments conducted on three popular medical image datasets reveal that our proposed model outperforms other existing state-of-the-art (SOTA) models and exhibits significant performance improvements. 
\end{itemize}
\section{Related Work} \label{sec:relatedwork}
\subsection{Medical Image Segmentation} \label{subsec:MIS}
With the development of Convolutional Neural Networks (CNNs), deep learning methods have become the dominated approach to MIS on various tumor \cite{tumor}, anatomical structures \cite{HyperDense-Net} and lesions \cite{H-DenseUNet} recently. Fully Convolutional Networks (FCNs), which replace the fully connected layers of standard CNNs with fully convolutional layers, prove to be a powerful net architecture for semantic segmentation \cite{FCNs}. Afterwards, the encoder-decoder networks gradually become the prevailing architecture for semantic segmentation, among which U-Net is a widely recognized method for MIS with remarkable results \cite{U-Net}. Besides, most recent MIS networks are designed based on visual transformer \cite{Swin-unet}, multi-scale fusion \cite{Vessel-Net}, feature pyramid \cite{DRINet} and self-attention mechanism \cite{CA-Net}. Unfortunately, all these models require abundant densely annotated data. 

    \begin{figure*}[t]
		\centering
		\includegraphics[width=0.99\textwidth]{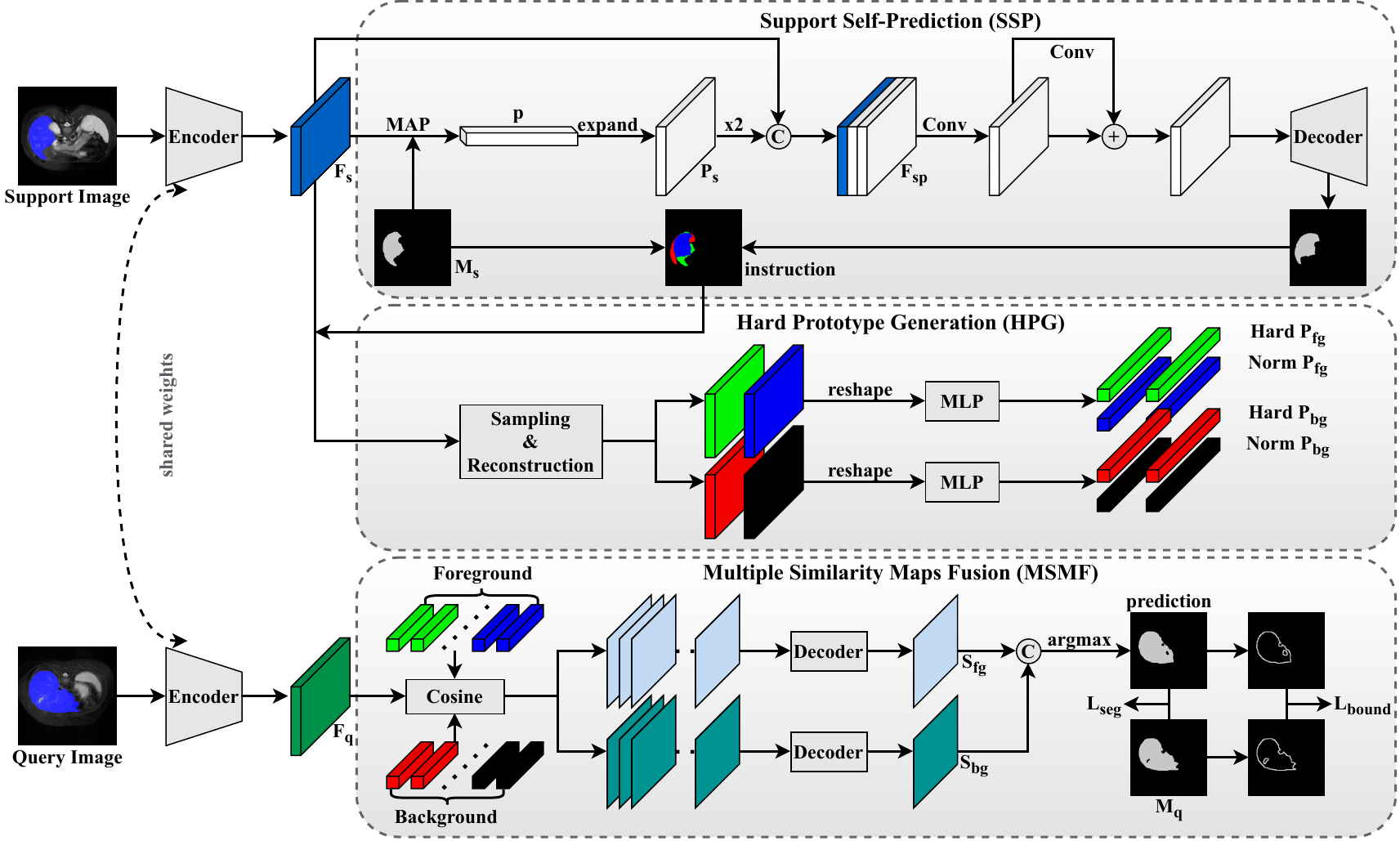}
		\caption{Illustration of CoW. We employ a shared feature encoder to learn deep features for both support and query images. In each episode, we firstly conduct SSP with support prototype $\smallp$ obtained via MAP. Next we compare this prediction with the ground truth $M_s$ to acquire the instruction for hard and normal feature points. Then we perform random sampling and reconstruct support feature map, which is used for generating multiple hard and normal prototypes for foreground and background, respectively. At last, these prototypes are applied to generate multiple prediction similarity maps that are fused by a lightweight decoder to get final segmentation mask.}
		\label{fig_method}
	\end{figure*}

\subsection{Few-Shot Medical Image Segmentation} \label{subsec:FSMIS}
FSMIS is a promising approach since it can work with a few labeled samples. The prototypical network \cite{PANet,ALPNet}, which extracts representative prototypes of semantic classes from the support, is one of prevalent FSMIS methods featured by its strong interpretability. PANet \cite{PANet} argues that parametric segmentation generalizes poorly and proposes a non-parametric prototype matching method with a prototype alignment regularization. ADNet \cite{ADNet} builds further on the branch initiated by PANet \cite{PANet} and does not explicitly model the complex background class namely relying solely on one foreground prototype. Q-Net \cite{QNet} replaces fixed thresholds used in ADNet with adaptive ones learned from the query to mitigate distribution shifts between the query image and the support set, and simultaneously predicts the query segmentation masks with dual-scale features. Most of these methods solely employ masked average pooling to extract a single prototype, which can not well represent the distribution of a specific category.

Unlike these single prototype-based methods, SSL-ALPNet \cite{ALPNet} devises an adaptive local prototype pooling module where additional local prototypes are computed on a regular grid to preserve local information lost in global average operation. AAS-DCL \cite{DCLAAS} constructs prototypical contrastive learning and contextual contrastive learning among the labeled support guidance information and unlabeled tissue knowledge, to strengthen the discriminability of the prototypes. GMRD \cite{GMRD} randomly selects multiple prototypes from both foreground and background to comprehensively represent the commonality within the corresponding class distribution. However, these methods pay no particular attention to weak boundary features even though they have constructed relatively comprehensive representation of each class with multiple prototypes.
\section{Methodology} \label{sec:method}
\subsection{Problem Definition} \label{subsec:definition}
The FSMIS task aims at training a model on a dataset $D_{train}$ that contains labeled samples belonging to seen classes $C_{base}$, allowing a quick adaptation to segmentation of previously unseen novel classes $C_{novel}$ when exposed to just a few annotated samples without retraining, where $C_{base}$ and $C_{novel}$ are disjoint, namely $C_{base} \cap C_{novel} = \varnothing$. In general, the episodic \cite{episode1,episode2} meta-learning setting is commonly utilized in FSS tasks. Specifically, by means of random sampling, the training set and the testing set are both divided into a support set and a query set (e.g., $D_{train} = \{(S_i, Q_i)\}_{i=1}^{N_{train}}$ and $D_{test} = \{(S_i, Q_i)\}_{i=1}^{N_{test}}$), where $N_{train}$ and $N_{test}$ are the number of episodes for training and testing respectively. When considering a N-way K-shot task, the support set $S_i = \{(I_s^k, M_s^k(c_j))\}_{k=1}^K$ contains $K$ image-mask pairs of an image $I \in \mathbb{R}^{H\times W}$ and its corresponding binary mask $M \in \{0, 1\}^{H\times W}$ for class $c_j \in C_{base}, j = 1,2,...,N$ and the query set $Q_i = \{(I_q^k, M_q^k(c_j))\}_{k=1}^{N_{qry}}$ contains $N_{qry}$ image-mask pairs from the same class as the support set. Notably, the background class denoted as $c_0$ does not count toward $C_{base}$ or $C_{novel}$. Given a support image-mask pair and a query image as input, the model for FSS task outputs predicted segmentation mask $\overset{\sim}{M}_q$ for the query image. 

\subsection{Network Overview} \label{subsec:overview}
The proposed CoW is a dual-path architecture where similar operations are performed, one for segmenting foreground regions and the other for background. The framework comprises following primary components: (1) a generic feature encoder shared by the support and query images for extracting features; (2) an \textbf{SSP} module for acquiring a coarse predicted mask of the support image using global support prototype; (3) an \textbf{HPG} module for generating a specified number of hard prototypes for both foreground and background; (4) an \textbf{MSMF} module for fusing multiple similarity maps to get final segmenting mask of the query image. Figure \ref{fig_method} shows a more detailed pipeline of our method.

\subsection{Support Self-Prediction} \label{subsec:suppselfpred}
As shown in Figure \ref{fig_method}, our proposed CoW starts by a parameter-sharing feature encoder denoted as $f_{\theta}$, which extracts features from both support and query images. We denote the support and the query features as $F_s = f_{\theta}(I_s)$ and $F_q = f_{\theta}(I_q), F_s, F_q \in \mathbb{R}^{C'\times H'\times W'}$, where $C'$ is the channel depth of the feature, and $H'$ and $ W'$ denote the height and width of the feature respectively. Following the practice in Q-Net \cite{QNet}, we adopt ResNet-101 pre-trained on MS-COCO \cite{mscoco} as the backbone, and utilize the output feature of the third residual block.

In order to find out the weak feature points easily misclassified, which lead to over-segmentation and boundary-blurring, we devise a Support Self-Prediction (SSP) module to provide clues for them. Specifically, we firstly employ masked average pooling to generate a global support prototype $\smallp_s \in \mathbb{R}^{1 \times C'}$ from the support feature map $F_s$:
\begin{equation} \label{eq1}
    \smallp_s = \frac{\sum_{h,w}F_s(h,w)\odot M_s(h,w)}{\sum_{h,w}M_s(h,w)},
\end{equation}
where $F_s$ is resized to the size of corresponding mask $M_s$. $\odot$ denotes the Hadamard product, and $(h,w)$ is the pixel position. Then we duplicate and expand the support prototype $\smallp_s$ to the same size as $F_s$, represented as $F_p$, which is subsequently concatenated with $F_s$ to generate a new feature map:
\begin{equation} \label{eq2}
    F_{sp} = Concat([F_s,F_p,F_p]),
\end{equation}
where $Concat(\cdot)$ denotes the concatenation operator.

Next, the similarity map for the support image is generated after passing through the feature processing module and the decoder that adopts single-scale residual layers following an ASPP module:
\begin{equation} \label{eq3}
    S_{support} = \mathcal{D}(FPM(F_{sp});\psi),
\end{equation}
where $S_{support} \in \mathbb{R}^{1\times H'\times W'}$ is the predicted support similarity map. $FPM(\cdot)$ is the feature processing operation and $\mathcal{D}(\cdot;\psi)$ means the support decoder parameterized with $\psi$. After that, we generate the predicted support mask $\hat{M}_s$ through a concatenation operation followed by an argmax operation:
\begin{equation} \label{eq4}
    \hat{M}_s = argmax(Concat([1-S_{support},S_{support}])),
\end{equation}
where $S_{support}$ is upsampled to $(H,W)$ in advance and $1-S_{support}$ represents the background similarity map.

For the purpose of finding weaker feature points in the support feature map, we adopt a cross-entropy loss on SSP module, which is written as:
\begin{equation} \label{eq5}
    \mathcal{L}_{ssp} = -\frac{1}{HW}\sum_{h,w}M_s^{bg}\logg(\hat{M}_s^{bg})+M_s^{fg}\logg(\hat{M}_s^{fg}),
\end{equation}
where $\hat{M}_s^{bg}=1-\hat{M}_s^{fg}$, and the superscript $fg$ and $bg$ is the indicator for the foreground and background respectively.

Finally, we calculate the position of weak feature points and normal ones borrowing information from both predicted support mask $\hat{M}_s$ and support ground truth $M_s$. This process can be mathematically expressed as:
\begin{equation} \label{eq6}
\left\{
\begin{array}{rcl}
    M_s^{hf} = [M_s(h,w)=1] \& [\hat{M}_s(h,w)=0], \\
    M_s^{hb} = [M_s(h,w)=0] \& [\hat{M}_s(h,w)=1], \\
    M_s^{nf} = [M_s(h,w)=1] \& [\hat{M}_s(h,w)=1], \\
    M_s^{nb} = [M_s(h,w)=0] \& [\hat{M}_s(h,w)=0],
\end{array}\right.
\end{equation}
where the superscript $hf,hb,nf,nb$ refers to hard foreground, hard background, normal foreground and normal background respectively.

\subsection{Hard Prototypes Generation} \label{subsec:HPG}
As opposed to previous works that employ multiple normal prototypes, which inevitably loses part of the detailed information and inadequately represents the overall feature distribution within a class, we devise a novel Hard Prototypes Generation (HPG) module to generate multiple hard prototypes for both foreground and background to guarantee full use of the support information. We will take the example of generating hard foreground prototypes as described below.

Building on the weak feature points instruction provided by the SSP module, we firstly remove all support feature points except for those belonging to hard foreground, which can be described as:
\begin{equation} \label{eq7}
    F_s^{hf} = F_s \odot M_s^{hf},
\end{equation}
where $\odot$ denotes element-wise multiplication.

Next, we reshape the features of hard foreground region, denoted as $D_{hf} = [d_{hf}^1,d_{hf}^2,...,d_{hf}^{N_{hf}}] \in \mathbb{R}^{N_{hf}\times C'}$. Then, a random sampling operation is performed on $D_{hf}$ to extract some hard foreground points $D_{hf}'$. This process can be expressed as:
\begin{equation} \label{eq8}
    D_{hf}' = \mathcal{F}(D_{hf},N),
\end{equation}
where $\mathcal{F}(\cdot)$ is a random sampling function and $N$ is the number of $D_{hf}'$.

Subsequently, we populate the hard foreground feature map $F_s^{hf}$ with $D_{hf}'$ where $M_s^{hf}=0$ and retain original weak foreground feature points, which can be summarized mathematically in the following formulation:
\begin{equation} \label{eq9}
F_s^{hf}(h,w)=\left\{
\begin{array}{rcl}
    &F_s^{hf}(h,w) &, M_s^{hf}(h,w)=1, \\
    &D_{hf}' &, M_s^{hf}(h,w)=0,
\end{array}\right.
\end{equation}
where $(h,w)$ denotes the spatial position of the pixel.

After that, we acquire a reconstructed hard foreground feature map where all feature points exclusively pertain to the hard foreground. In order to generate multiple hard foreground prototypes that fully capture the feature information, we pass the reconstructed $F_s^{hf}$ through a lightweight MLP which is composed of a fully connected layer, a ReLU activation layer, and another fully connected layer. This process can be denoted as:
\begin{equation} \label{eq10}
    \bigp_{hf} = MLP(F_s^{hf};\phi),
\end{equation}
where $MLP(\cdot;\phi)$ represents the MLP module parameterized with $\phi$. The hard foreground feature map is reshaped to the dimension of $HW\times C'$ before being fed into the MLP and $\bigp_{hf} \in \mathbb{R}^{N_{hf}\times C'}$ denotes generated $N_{hf}$ hard foreground prototypes.

Notably, the aforementioned procedure for generating hard prototypes also applies to hard background, normal foreground and background, namely $\bigp_{hb},\bigp_{nf},\bigp_{nb}$. Eventually, we concatenate the generated hard prototypes with normal ones and corresponding global support prototype for foreground and background respectively, which can be formulated as:
\begin{equation} \label{eq11}
\left\{
\begin{array}{rcl}
\bigp_{fg} =& \bigp_{hf}\oplus \bigp_{nf}\oplus \smallp_s^{fg},  \\
\bigp_{bg} =& \bigp_{hb}\oplus \bigp_{nb}\oplus \smallp_s^{bg},
\end{array}\right.
\end{equation}
where $\oplus$ denotes the concatenation operation. $\smallp_s^{fg}$ and $\smallp_s^{bg}$ are support foreground and background prototypes generated via MAP. $\bigp_{fg} \in \mathbb{R}^{N_{fg}\times C'}$ and $\bigp_{bg} \in \mathbb{R}^{N_{bg}\times C'}$ are final representative prototypes used for predicting the foreground and background similarity maps of the query image respectively.

\subsection{Multiple Similarity Maps Fusion} \label{subsec:MSMF}
In previous FSMIS tasks, researchers commonly leverage a single prototype or multiple normal prototypes to make predictions for the query image directly based on cosine-similarity. However, there exist extreme imbalance between foreground and background in medical image and lost information in the masked feature, which easily lead to misclassified foreground and background points. To tackle this challenge, we design a Multiple Similarity Maps Fusion (MSMF) module, which fuses similarity maps generated by multiple prototypes obtained from HPG module, thus making more accurate segmentation.

First of all, we calculate the similarity maps between the multiple foreground and background prototypes, and the query image respectively:
\begin{equation} \label{eq12}
\left\{
\begin{array}{rcl}
    S_{fg}(h,w) = \frac{\bigp_{fg}^T \cdot F_q(h,w)}{\left\| \bigp_{fg} \right\| \cdot \left\| F_q(h,w) \right\|},  \\
    S_{bg}(h,w) = \frac{\bigp_{bg}^T \cdot F_q(h,w)}{\left\| \bigp_{bg} \right\| \cdot \left\| F_q(h,w) \right\|},
\end{array}\right.
\end{equation}
where $S_{fg} \in \mathbb{R}^{N_{fg}\times H'\times W'}$ and $S_{bg} \in \mathbb{R}^{N_{bg}\times H'\times W'}$ denote the generated similarity maps for foreground and background of the query respectively.

Then, both $S_{fg}$ and $S_{bg}$ will be fed into a lightweight decoder to fuse multiple similarity maps, thereby generating the foreground prediction map and background prediction map. This process can be denoted as:
\begin{equation} \label{eq13}
\left\{
\begin{array}{rcl}
    \hat{M}_q^{fg} = \mathcal{D}_1(S_{fg};\psi_1), \\
    \hat{M}_q^{bg} = \mathcal{D}_2(S_{bg};\psi_2),
\end{array}\right.
\end{equation}
where $\mathcal{D}_1(\cdot;\psi_1)$ and $\mathcal{D}_2(\cdot;\psi_2)$ denote a straightforward decoder based on convolutional neural network with parameter $\psi_1$ and $\psi_2$. $\hat{M}_q^{fg} \in \mathbb{R}^{1\times H'\times W'}$ and $\hat{M}_q^{bg} \in \mathbb{R}^{1\times H'\times W'}$ are the final single predicted similarity map for query foreground and background respectively.

Finally, we upsample $\hat{M}_q^{fg}$ and $\hat{M}_q^{bg}$ to the size of $(H,W)$ and integrate them by utilizing the concatenation operation and a $\softmax$ layer to get the final prediction, as follows:
\begin{equation} \label{eq14}
    \hat{M}_q = softmax(\hat{M}_q^{bg}\oplus \hat{M}_q^{fg}),
\end{equation}
where $softmax(\cdot)$ executes along the channel dimension and $\hat{M}_q \in \mathbb{R}^{2\times H\times W }$ will eventually be processed by an $\argmax$ operation to get the final segmentation mask of the query image.
\begin{table*}[!t]
    \centering
    \resizebox{\textwidth}{!}{
    \setlength{\tabcolsep}{3mm}
    \begin{tabular}{c|c|c|ccccc|ccccc}
        \toprule
        
        \multirow{2}{*}{Setting} & \multirow{2}{*}{Method} & \multirow{2}{*}{Reference} & \multicolumn{5}{c|}{Abd-MRI} & \multicolumn{5}{c}{Abd-CT} \\
        & & & Liver & RK & LK & Spleen & Mean & Spleen & RK & LK & Liver & Mean \\
        
        \hline
        
        \multirow{9}{*}{1} & PA-Net & CVPR'19 & 50.40 & 32.19 & 30.99 & 40.58 & 38.53 & 36.04 & 21.19 & 20.67 & 49.55 & 31.86 \\
        & SSL-ALPNet & ECCV'20 & 76.10 & 85.18 & 81.92 & 72.18 & 78.84 & 70.96 & 71.81 & 72.36 & 78.29 & 73.35 \\
        & SR\&CL & MICCAI'22 & 80.23 & 87.42 & 79.34 & 76.01 & 80.77 & 73.41 & 71.22 & 73.45 & 76.06 & 73.53 \\
        & Q-Net & IntelliSys'23 & \underline{81.74} & 87.98 & 78.36 & 75.99 & 81.02 & 77.81 & 70.98 & 73.83 & 76.37 & 74.75 \\
        & CAT-Net & MICCAI'23 & 75.02 & 83.23 & 75.31 & 67.31 & 75.22 & 66.02 & 64.56 & 68.82 & 80.51 & 70.88 \\
        & RPT & MICCAI'23 & \textbf{82.86} & 89.82 & 80.72 & 76.37 & 82.44 & \underline{79.13} & 72.58 & 77.05 & \textbf{82.57} & 77.83 \\
        & GMRD & TMI'24 & 81.42 & \underline{90.12} & 83.96 & 76.09 & \underline{82.90} & 78.31 & 74.46 & \underline{81.70} & 79.60 & \underline{78.52} \\
        & DSPNet & MIA'24 & 75.06 & 85.37 & 81.88 & 70.93 & 78.31 & 69.31 & 74.54 & 78.01 & 69.32 & 72.79 \\
        & MSFSeg & MICCAI'24 & 76.11 & 88.10 & \underline{84.18} & \underline{77.12} & 81.38 & 73.64 & \underline{78.41} & 81.11 & 78.91 & 78.02 \\
        & \textbf{Ours (CoW)} & -- & 81.55 & \textbf{90.27} & \textbf{85.58} & \textbf{79.00} & \textbf{84.10} & \textbf{83.56} & \textbf{80.46} & \textbf{83.60} & \underline{82.35} & \textbf{82.49} \\
        
        \hline
        
        \multirow{9}{*}{2} & PA-Net & CVPR'19 & 42.26 & 38.64 & 53.45 & 50.90 & 46.33 & 29.59 & 17.37 & 32.34 & 38.42 & 29.43 \\
        & SSL-ALPNet & ECCV'20 & 73.05 & 78.39 & 73.63 & 67.02 & 73.02 & 60.25 & 54.82 & 63.34 & 73.65 & 63.02 \\
        & SR\&CL & MICCAI'22 & 75.55 & 84.24 & 77.07 & 73.73 & 77.65 & 67.36 & 63.37 & 67.39 & 73.63 & 67.94 \\
        & Q-Net & IntelliSys'23 & 78.25 & 65.94 & 64.81 & 65.37 & 68.59 & 68.10 & 61.12 & 63.32 & \underline{80.49} & 68.26 \\
        & CAT-Net & MICCAI'23 & 78.98 & 78.90 & 74.01 & 68.83 & 75.18 & 67.65 & 60.05 & 63.36 & 75.31 & 66.59 \\
        & RPT & MICCAI'23 & 76.37 & 86.01 & 78.33 & \underline{75.46} & 79.04 & 70.80 & 67.73 & 72.99 & 75.24 & 71.69 \\
        & GMRD & TMI'24 & \underline{80.25} & 86.66 & 78.65 & 73.25 & 79.70 & \underline{75.30} & 76.17 & 77.40 & 80.39 & \underline{77.32} \\
        & DSPNet & MIA'24 & 78.56 & 82.01 & 76.47 & 68.27 & 76.33 & 66.48 & 63.55 & 68.46 & 69.16 & 66.17 \\
        & MSFSeg & MICCAI'24 & 76.14 & \underline{86.98} & \textbf{82.83} & \textbf{78.07} & \underline{81.01} & 75.21 & \underline{77.36} & \underline{79.24} & 76.73 & 77.14 \\
        & \textbf{Ours (CoW)} & -- & \textbf{80.46} & \textbf{89.15} & \underline{82.40} & 74.55 & \textbf{81.64} & \textbf{79.92} & \textbf{79.44} & \textbf{83.36} & \textbf{82.26} & \textbf{81.25} \\
        
        \bottomrule
    \end{tabular}}
    \caption{Quantitative comparison (in DSC score \%) of different methods under Setting 1 and Setting 2 on Abd-MRI and Abd-CT datasets. Bold and underlined numbers denote the best and second best results, respectively.}
    \label{table1}
\end{table*}
\subsection{Loss Function} \label{subsec:loss}
In order to quantify the dissimilarity between the predicted and ground truth, we employ the binary cross-entropy loss as the segmentation loss for each training episode:
\begin{equation} \label{eq15}
    \mathcal{L}_{seg} = -\frac{1}{HW}\sum_{h,w}M_q^{bg}\logg(\hat{M}_q^{bg})+M_q^{fg}\logg(\hat{M}_q^{fg}),
\end{equation}
where $\mathcal{L}_{seg}$ works as the primary loss of our proposed CoW.

Following common practice \cite{PANet,ALPNet}, we inversely predict labels of the support images by using query images as the support set, then build a prototype alignment regularization loss, which is calculated as follows:
\begin{equation} \label{eq16}
    \mathcal{L}_{align} = -\frac{1}{HW}\sum_{h,w}M_s^{bg}\logg(\hat{M}_s^{bg})+M_s^{fg}\logg(\hat{M}_s^{fg}).
\end{equation}

Additionally, we employ the intra-class loss to help diminish intra-class variation of the support and query and the inter-class loss to improve discrimination among different classes, both enhancing the representativeness of our generated hard and normal prototypes for foreground and background, which is formulated as:
\begin{align} 
    &\mathcal{L}_{intra}=2-\sum_{*\in\{bg,fg\}}\frac{1}{N_*}\sum_{k=1}^{N_*}\max_{\smallp_k^s\in\bigp_*^s}(\cos(\smallp_k^s,\bigp_*^q)), \\
    &\mathcal{L}_{inter}=\sum_{*\in\{s,q\}}\frac{1}{N_{fg}N_{bg}}\sum_{\smallp_i^*\in\bigp_{fg}^*}\sum_{\smallp_j^*\in\bigp_{bg}^*}\cos(\smallp_i^*,\smallp_j^*),
\end{align}
where the superscripts $s$ and $q$ denote the support and query image, $\cos$ is the cosine similarity, and $\max(\cdot)$ means taking the maximum.

Furthermore, to get a more precise predicted mask, we devise a boundary loss to refine the segmentation boundary. Specifically, we generate the boundary map of ground truth and predicted mask through a $maxpool$ operation, denoted as $M_q^{b}$ and $\hat{M}_q^{b}$. Then, we calculate f1-score between them and get final boundary loss $\mathcal{L}_{bound}$ as follows:
\begin{equation} \label{eq19}
    \mathcal{L}_{bound} = 1-BF1(M_q^{b},\hat{M}_q^{b}),
\end{equation}
where $BF1(\cdot,\cdot)$ computes the f1-score between input pair.

Overall, the loss function for each training episode is defined to be $\mathcal{L} = \mathcal{L}_{seg} + \mathcal{L}_{align} + \lambda_0\mathcal{L}_{bound} + \lambda_1(\mathcal{L}_{intra} + \mathcal{L}_{inter} + \mathcal{L}_{ssp})$, where $\lambda_0,\lambda_1$ are the balancing coefficients.
\section{Experiment} \label{sec:experiment}
\subsection{Datasets, Settings and Evaluation Metrics} \label{subsec:datasets}

\paragraph{Datasets.} To demonstrate the superiority of our method under different scenarios, we perform evaluations on three representative publicly available datasets: abdominal organs segmentation for MRI and CT (Abd-MRI \cite{Abd-MRI} and Abd-CT \cite{Abd-CT}) and cardiac segmentation for MRI (CMR \cite{CMR}). The detailed introduction and split strategy is depicted in \cite{RPT}.

\paragraph{Settings.} The majority of FSMIS models follow the settings in the work \cite{ssl} to evaluate their performance. In order to make a fair comparison, we also adopt the two experimental settings outlined in SSL-ALPNet \cite{ALPNet} and ADNet \cite{ADNet}.

\paragraph{Evaluation Metrics.} Following the common practice \cite{SENet} in FSMIS, we employ the mean Sorensen-Dice coefficient (DSC) to compare the prediction to the ground truth. It measures the effectiveness of the same area between the predicted mask and the ground-truth.

\subsection{Implementation Details}
Following the practice in SSL-ALPNet \cite{ALPNet}, the experiments are conducted based on a 1-way 1-shot setting with 5-fold cross-validation. Initially, pseudo labels are generated for training by leveraging 3D superpixel clustering. After that, we reshape the size of the 2D slice from the 3D scanned image to $256\times256$ for Abd-MRI and CMR and $257\times257$ for Abd-CT. The total number of generated hard and normal prototypes for foreground and background are set to $100$ and $600$ respectively. We set the initial learning rate to $0.001$, the decay rate to $0.9$, and the total number of required iterations to $50K$, containing $5000$ iterations each epoch. The balancing coefficients $\lambda_0$ and $\lambda_1$ are set to $0.5$ and $0.3$ respectively in our experiment.

\begin{table}[!t]
    \centering    
    \resizebox{0.48\textwidth}{!}{
    \begin{tabular}{c|c|cccc}
        \toprule        
        \multirow{2}{*}{Method} & \multirow{2}{*}{Reference} & \multicolumn{4}{c}{CMR} \\
        & & LV-MYO & LV-BP & RV & Mean \\
        \hline
        PA-Net & CVPR'19 & 25.18 & 58.04 & 12.86 & 32.02 \\
        SSL-ALPNet & ECCV'20 & 66.74 & 83.99 & 79.96 & 76.90 \\
        SR\&CL & MICCAI'22 & 65.83 & 84.74 & 78.41 & 76.32 \\
        Q-Net & IntelliSys'23 & 65.92 & \underline{90.25} & 78.19 & 78.15 \\
        CAT-Net & MICCAI'23 & 66.85 & \textbf{90.54} & 79.71 & 79.03 \\
        RPT & MICCAI'23 & 66.91 & 89.90 & \underline{80.78} & \underline{79.19} \\
        GMRD & TMI'24 & \underline{67.04} & 90.00 & 80.29 & 79.11 \\
        DSPNet & MIA'24 & 64.91 & 87.75 & 79.73 & 77.46 \\
        \textbf{Ours (CoW)} & -- & \textbf{68.41} & 89.25 & \textbf{82.33} & \textbf{80.00} \\
        \bottomrule        
    \end{tabular}}
    \caption{Quantitative comparison (in DSC score \%) of different methods under Setting 1 on CMR dataset. Bold and underlined numbers denote the best and second best results, respectively.}
    \label{table2}
\end{table}

\subsection{Comparison with SOTA Methods} \label{subsec:results}
For the purpose of convincingly demonstrating the superiority and effectiveness of our proposed CoW, we compare our method with several existing state-of-the-art methods, including PA-Net \cite{PANet}, SSL-ALPNet \cite{ALPNet}, SR\&CL \cite{SRCL}, CAT-Net \cite{CATNet}, Q-Net \cite{QNet}, RPT \cite{RPT}, GMRD \cite{GMRD}, DSPNet \cite{DSPNet} and MSFSeg \cite{MSFSeg}, from the perspective of segmentation performance under two experimental settings. In this paper, considering that most researchers commonly adopt the PA-Net \cite{PANet} as the baseline and the settings in SSL-ALPNet \cite{ALPNet} to evaluate their performance, we also follow them to ensure a fair comparison. As shown in Table \ref{table1} and Table \ref{table2}, our CoW significantly outperforms all listed methods in terms of the mean dice score on three different datasets.

\begin{table}[!t]
    \centering       
    \resizebox{0.48\textwidth}{!}{
    \begin{tabular}{cccc|ccccc}
        \toprule
        
        \multirow{2}{*}{Baseline} & \multirow{2}{*}{SSP} & \multirow{2}{*}{HPG} & \multirow{2}{*}{MSMF}& \multicolumn{5}{c}{Abd-MRI} \\
        & & & & Liver & RK & LK & Spleen & Mean \\

        \hline
        \Checkmark & \XSolidBrush & \XSolidBrush & \XSolidBrush & 50.40 & 32.19 & 30.99 & 40.58 & 38.54 \\
        \Checkmark & \Checkmark & \Checkmark & \XSolidBrush & 80.29 & 88.57 & 81.38 & 75.52 & 81.44 \\
        \Checkmark & \Checkmark & \Checkmark & \Checkmark & \textbf{81.55} & \textbf{90.27} & \textbf{85.58} & \textbf{79.00} & \textbf{84.10} \\
        
        \bottomrule
        
    \end{tabular}}
    \caption{Ablation study on Abd-MRI dataset under Setting 1 for the effect of each component on the performance of CoW in terms of the DICE score.}
    \label{table3}
\end{table}

\begin{table}[!t]
    \centering    
    \resizebox{0.48\textwidth}{!}{
    \begin{tabular}{cccc|ccccc}
        \toprule
        
        \multirow{2}{*}{$\mathcal{L}_{intra}$} & \multirow{2}{*}{$\mathcal{L}_{inter}$} & \multirow{2}{*}{$\mathcal{L}_{bound}$} & \multirow{2}{*}{$\mathcal{L}_{ssp}$}& \multicolumn{5}{c}{Abd-MRI} \\
        & & & & Liver & RK & LK & Spleen & Mean \\

        \hline
        \XSolidBrush & \XSolidBrush & \XSolidBrush & \XSolidBrush & 80.77 & 89.12 & 77.68 & 73.57 & 80.29 \\
        \Checkmark & \XSolidBrush & \XSolidBrush & \XSolidBrush & 80.88 & 89.91 & 78.22 & 75.13 & 81.04 \\
        \Checkmark & \Checkmark & \XSolidBrush & \XSolidBrush & 80.91 & 89.95 & 79.62 & 75.74 & 81.56 \\
        \Checkmark & \Checkmark & \Checkmark & \XSolidBrush & 81.16 & 90.22 & 83.91 & 78.27 & 83.39 \\
        \Checkmark & \Checkmark & \Checkmark & \Checkmark & \textbf{81.55} & \textbf{90.27} & \textbf{85.58} & \textbf{79.00} & \textbf{84.10} \\
        
        \bottomrule
        
    \end{tabular}}
    \caption{Ablation study on Abd-MRI dataset under Setting 1 for the effect of auxiliary losses on the performance of CoW.}
    \label{table4}
\end{table}

\subsection{Ablation Studies} \label{subsec:ablation}
\paragraph{Effect of each component.} To evaluate separate contributions of each component within our model, we present their positive contributions in Table \ref{table3}. Our SSP module and HPG module significantly improve its performance on top of the baseline model \cite{PANet}, achieving a mean dice score of 81.44\%, which is attributed to the combination of normal and hard prototypes carrying rich subject and boundary information. In addition, the MSMF module with a dual-path architecture that aims at fusing multiple similarity maps further improves the performance by 2.66\%.

\begin{table}[!t]
    \centering    
    \begin{tabular}{cc|ccccc}
        \toprule
        
        \multirow{2}{*}{$N_{hf}$} & \multirow{2}{*}{$N_{nf}$} & \multicolumn{5}{c}{Abd-MRI} \\
        & & Liver & RK & LK & Spleen & Mean \\

        \hline
        0 & 100 & 81.49 & 89.22 & 84.09 & 77.44 & 83.06 \\
        25 & 75 & 81.55 & 90.03 & 84.39 & 78.66 & 83.66 \\
        50 & 50 & 81.55 & \textbf{90.27} & \textbf{85.58} & \textbf{79.00} & \textbf{84.10} \\
        75 & 25 & 81.22 & 89.85 & 82.99 & 77.83 & 82.97 \\
        100 & 0 & \textbf{82.04} & 90.15 & 83.75 & 72.43 & 82.09 \\
        
        \bottomrule
        
    \end{tabular}
    \caption{Ablation study on Abd-MRI dataset under Setting 1 for the effect of number of $\bigp_{hf}$ and $\bigp_{nf}$ on the performance of CoW, where $N_{hf} + N_{nf} = 100, N_{hb} = 100$ and $N_{nb} = 500$.}
    \label{table5}
\end{table}

\paragraph{Effect of each loss.} As can be seen from the last column in Table \ref{table4}, each loss indeed makes a separate contribution to the final performance. $\mathcal{L}_{intra}$ and $\mathcal{L}_{inter}$ achieve a total gain of 1.27\% by improving discriminability of generated multiple prototypes. Particularly, $\mathcal{L}_{bound}$ gives a 1.83\% increase due to its constraint on the segmentation boundary. Moreover, $\mathcal{L}_{ssp}$ brings an additional gain of 0.71\% for the reason that we can acquire more robust normal and hard prototypes making support self-prediction more precise. 

\paragraph{Number of normal and hard foreground prototypes.} We test the effect of the number for $\bigp_{hf}$ and $\bigp_{nf}$ on the final performance of CoW. As shown in Table \ref{table5}, the optimal number setting is half for each. By contrast, employing $\bigp_{hf}$ or $\bigp_{nf}$ alone leads to a performance degradation of 2.01\% and 1.04\% respectively because they solely leverage singular subject or boundary information, thus causing a less comprehensive representation for each category.

\paragraph{Generating multiple prototypes.} We compare our method with several previous works \cite{ALPNet,PMM,K-means,GMRD} in Table \ref{table6}. It is apparent that our CoW achieves the best performance, surpassing the suboptimal GMRD by 3.97\%, for the reason of focusing more attention to weak boundary details.

\begin{table}[!t]
    \centering  
    \resizebox{0.48\textwidth}{!}{
    \begin{tabular}{c|cccccc}
        \toprule
        
        \multirow{2}{*}{Method} & \multicolumn{6}{c}{Abd-CT} \\
        & Fold0 & Fold1 & Fold2 & Fold3 & Fold4 & Mean \\

        \hline
        SSL-ALPNet & 75.37 & 65.65 & 68.68 & 77.68 & 79.37 & 73.35 \\
        PMM & 74.70 & 63.81 & 69.33 & 78.26 & 76.15 & 72.45 \\
        K-means & 77.06 & 67.02 & 75.42 & 79.76 & 80.79 & 76.01 \\
        GMRD & 79.30 & 72.62 & 76.92 & 80.20 & 83.55 & 78.52 \\
        \textbf{Ours (CoW)} & \textbf{83.46} & \textbf{77.72} & \textbf{81.14} & \textbf{84.32} & \textbf{85.80} & \textbf{82.49} \\
        
        \bottomrule
        
    \end{tabular}}
    \caption{Ablation study on Abd-CT dataset under Setting 1 about different methods for generating multiple prototypes.}
    \label{table6}
\end{table}

\begin{figure}[!t]
    \centering
    \includegraphics[width=0.48\textwidth]{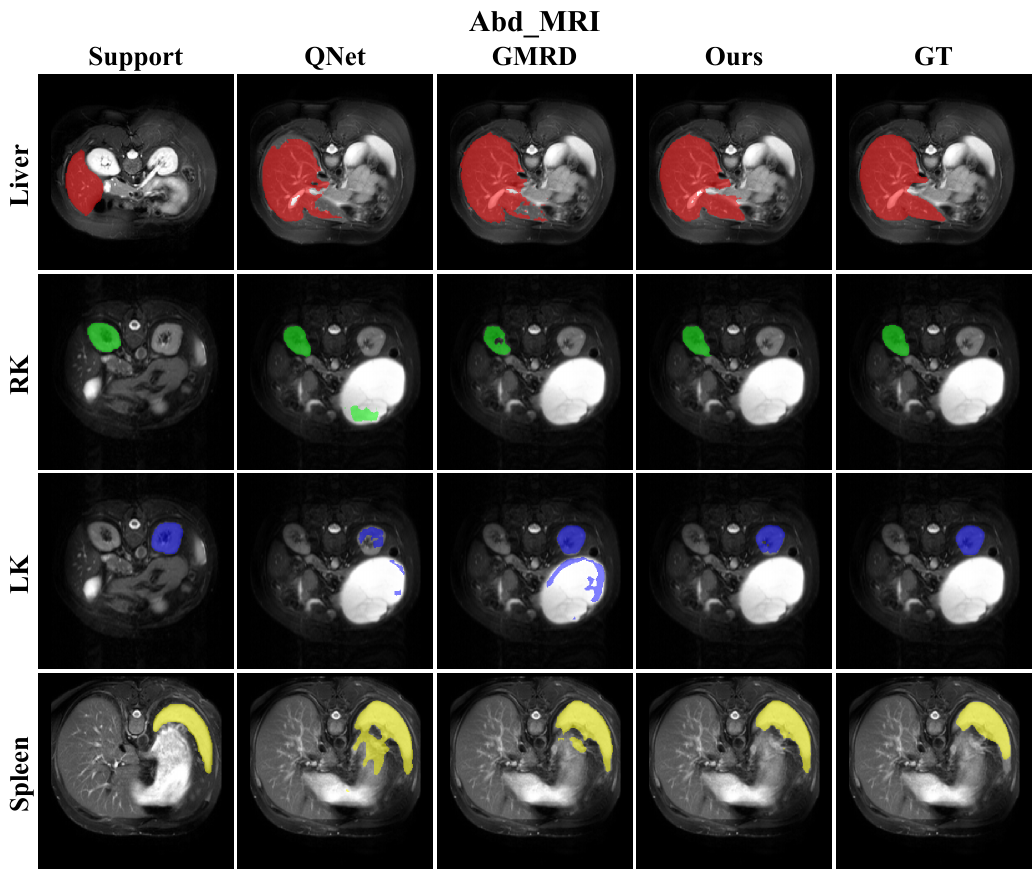}
    \caption{Comparison of qualitative results between our method and other on Abd-MRI.}
    \label{fig_abd_mri}
\end{figure}

\begin{figure}[!t]
    \centering
    \includegraphics[width=0.48\textwidth]{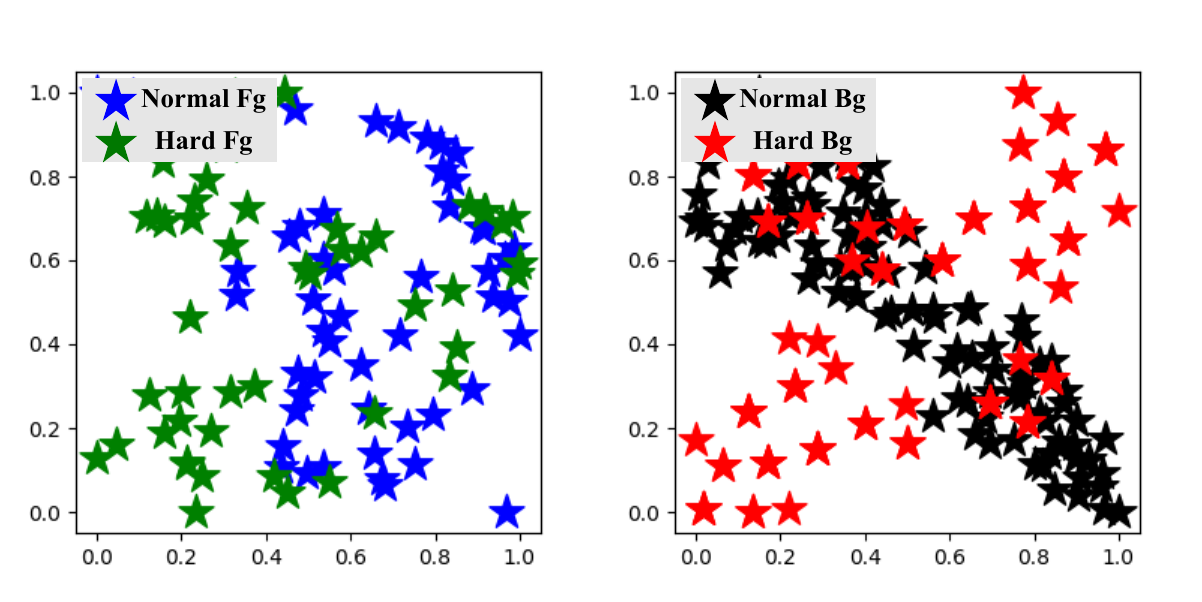}
    \caption{t-SNE visualization for generated normal and hard prototypes. `Fg' implies foreground and `Bg' implies background.}
    \label{fig_t-SNE}
\end{figure}

\subsection{Visualization}
\paragraph{Qualitative Results.} To better demonstrate the superiority of our method, qualitative results of our model and other models, including Q-Net \cite{QNet} and GMRD \cite{GMRD}, on Abd-MRI are shown in Fig. \ref{fig_abd_mri} (More qualitative results on the CMR dataset and the Abd-CT dataset can be found in Appendix). 

\paragraph{Distribution of Hard and Normal Prototypes.} As depicted in Fig. \ref{fig_t-SNE}, we visualize the generated hard and normal prototypes with t-SNE \cite{t-SNE}. It can be seen that normal foreground or background prototypes alone fail to well represent the entire distribution of their corresponding class. In contrast, the introduction of hard prototypes can compensate for such deficiency.

\section{Conclusion}
In this paper, we have proposed to mine hard prototypes by concentrating on weak features for few-shot medical image segmentation. It alleviates the challenge of current FSMIS methods that generate multiple prototypes solely through random sampling and local averaging, often failing to give a complete representation of corresponding category distribution. Additionally, our model includes a powerful edge constraint, which can further ensure a clearer segmentation boundary. Numerous experimental results also demonstrate that our method can capture more comprehensive class distribution and precisely segment the
intricate organs in medical images. We believe that our model has the potential to assist in the diagnosis of rare disease and improve the efficiency of medical treatment.
 
\section*{Acknowledgments}
This work was partly supported by the National Natural Science Foundation of China (NSFC) under Grant No. 62371235, and partly by the Key Research and Development Plan of Jiangsu Province (Industry Foresight and Key Core Technology Project) under Grant BE2023008-2.



\begin{center}
    \Large\bfseries Appendix\normalsize
\end{center}
In this section, we present the complete experimental results omitted from the main text, along with additional validation, analysis, and further details on the implementation of Hard Prototype Generation (HPG) module.

\paragraph{Algorithm of HPG.}To more clearly depict the HPG strategy, we provide a more detailed implementation algorithm table as follows.

\begin{algorithm}
\renewcommand{\algorithmicrequire}{\textbf{Input:}}
\renewcommand{\algorithmicensure}{\textbf{Output:}}
\caption{Process of Hard Prototype Generation (HPG)}
\label{alg:hpg}
\begin{algorithmic}[1]
\REQUIRE Support mask $M_s$ and support feature $F_s$.
\STATE Compute support prototype $\smallp_s$ via MAP (Eq. \ref{eq1}).
\STATE Duplicate and expand $\smallp_s$ to the size of $F_s$ to get $F_p$.
\STATE Generate $F_{sp}$ by concatenating $F_s$ with two $F_p$ (Eq. \ref{eq2}).
\STATE Process $F_{sp}$ and generate predicted support mask $\hat{M}_s$ (Eq. \ref{eq3}, \ref{eq4}).
\STATE Obtain hard foreground mask $M_s^{hf}$ by comparing $M_s$ and $\hat{M}_s$ (Eq. \ref{eq6}).
\STATE Remove all other feature points except for hard foreground to get $F_s^{hf}$ (Eq. \ref{eq7}).
\STATE Reshape $F_s^{hf}$ and randomly select $N$ hard foreground points $D_{hf}'$ (Eq. \ref{eq8}).
\STATE Reconstruct hard foreground feature map $F_s^{hf}$ with selected points $D_{hf}'$ (Eq. \ref{eq9}).
\IF{$M_s^{hf}(h,w)=1$}
\STATE Retain original weak foreground feature points.
\ELSE
\STATE Substitute feature points of $F_s^{hf}$ with those in $D_{hf}'$.
\ENDIF
\STATE Reshape reconstructed $F_s^{hf}$ and pass it through a $MLP$ module to generate hard foreground prototypes $\bigp_{hf}$ (Eq.\ref{eq10}).
\ENSURE Generated hard foreground prototypes $\bigp_{hf}$.
\end{algorithmic}
\end{algorithm}

\paragraph{Effect of different pre-trained datasets for backbone.} Table. \ref{table7} shows the effect of different pre-trained datasets on the model’s performance. Obviously, backbone pre-trained on MS-COCO \cite{mscoco} consistently outperforms that pre-trained on
ImageNet \cite{ImageNet} under both settings, which can be attributed to the fact that ImageNet is mainly applied for classification and MS-COCO is commonly used for segmentation.

\paragraph{Number of normal and hard background prototypes.} We test the effect of the number for $\bigp_{hb}$ and $\bigp_{nb}$ on the final performance of CoW. As shown in Table \ref{table8}, the optimal number setting is $100$ for $\bigp_{hb}$ and $500$ for $\bigp_{nb}$ . By comparison, employing $\bigp_{hb}$ or $\bigp_{nb}$ alone leads to a performance degradation of 2.67\% and 0.61\% respectively because they overlook subject or boundary information.

\begin{figure}[!t]
    \centering
    \includegraphics[width=0.48\textwidth]{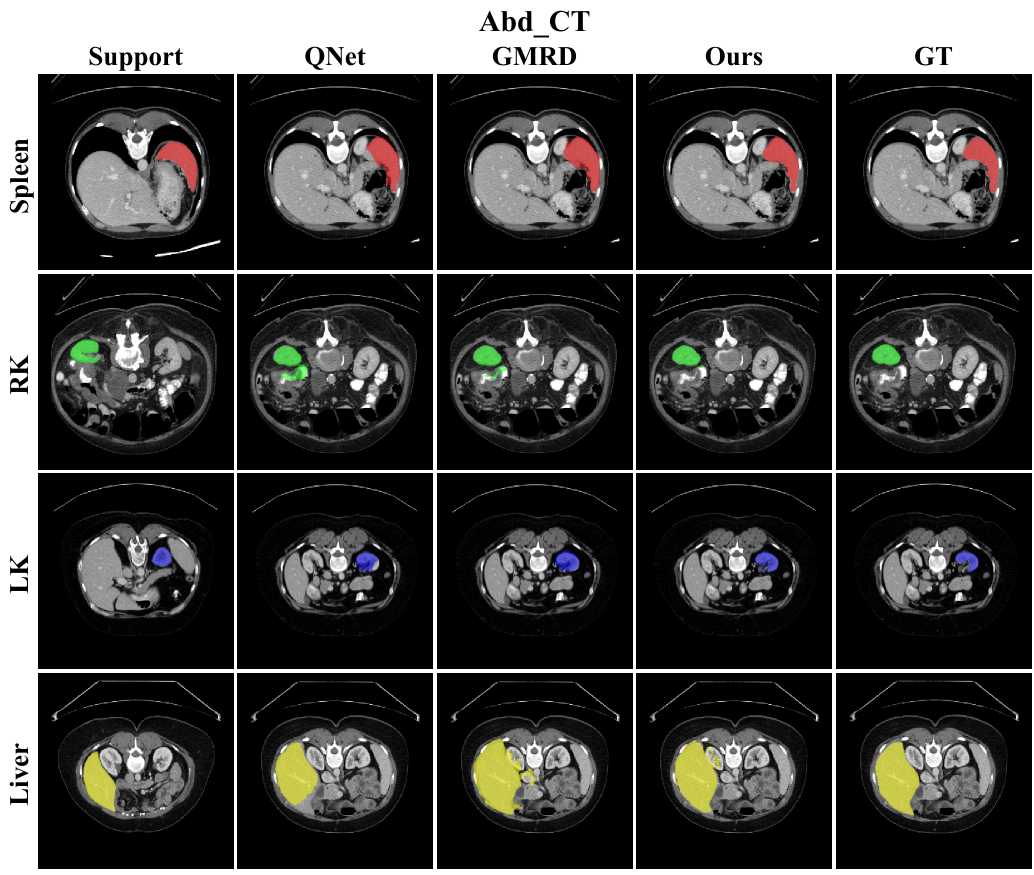}
    \vspace{-4ex}
    \caption{Comparison of qualitative results between our method and other on Abd-CT.}
    \label{fig_abd_ct}
    \vspace{-2ex}
\end{figure}

\begin{figure}[!t]
    \centering
    \includegraphics[width=0.48\textwidth]{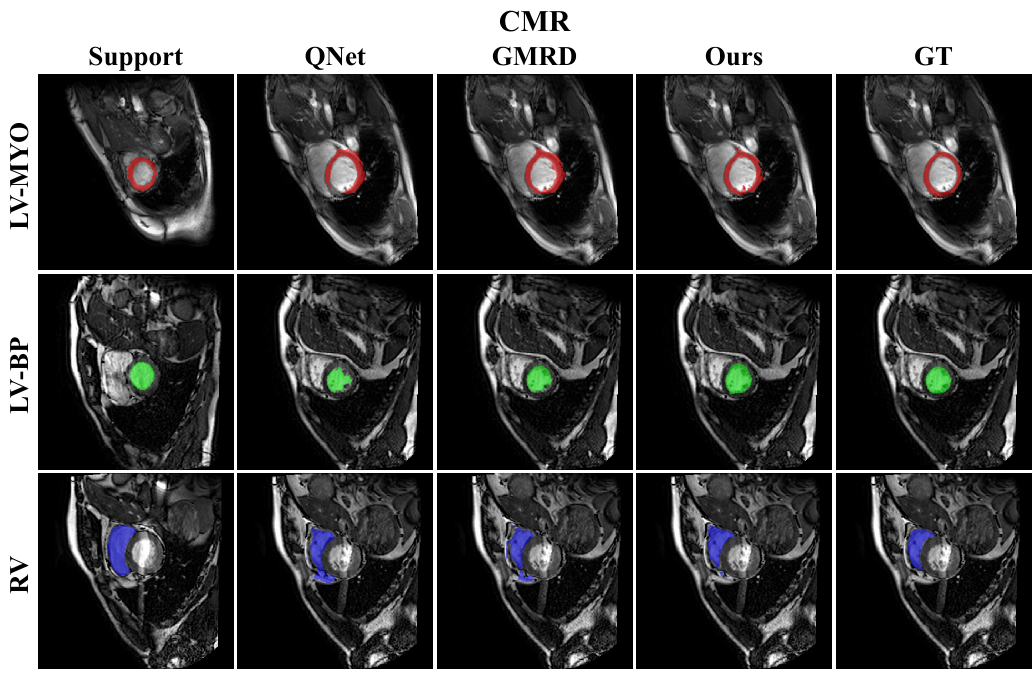}
    \caption{Comparison of qualitative results between our method and other on CMR.}
    \label{fig_cmr}
    \vspace{-2ex}
\end{figure}

\begin{figure*}[!t]
    \centering
    \includegraphics[width=0.75\textwidth]{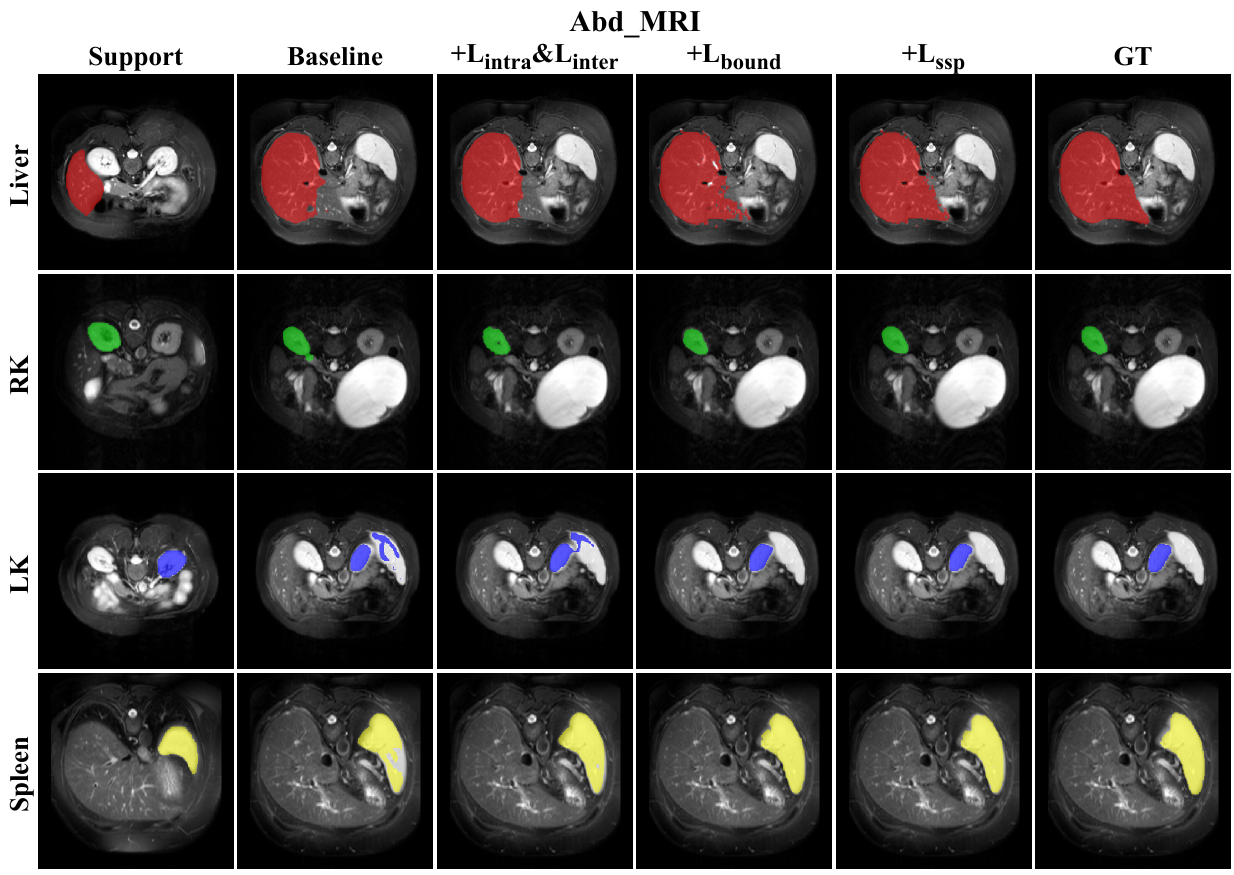}
    \caption{Ablation study on Abd-MRI dataset under setting 1 for the effect of each auxiliary loss.}
    \label{fig_loss-ablation}
    \vspace{-2ex}
\end{figure*}

\begin{figure*}[!t]
    \centering
    \includegraphics[width=0.75\textwidth]{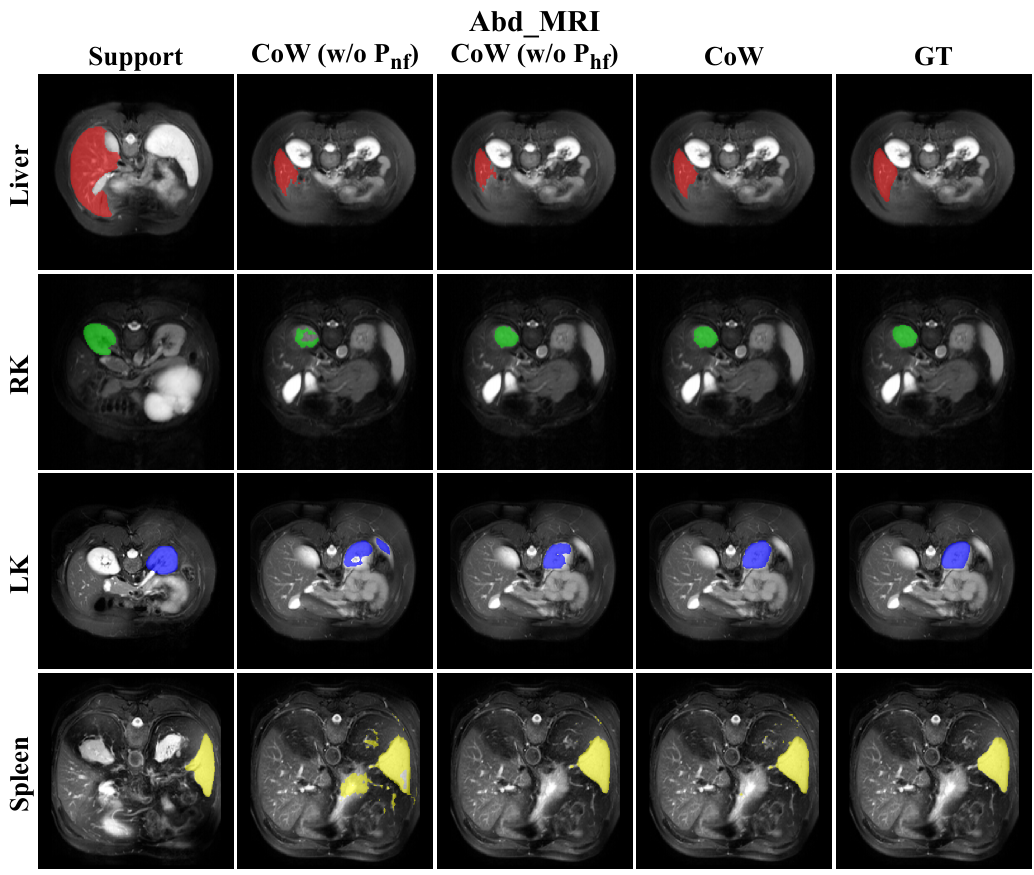}
    \caption{Ablation study on Abd-MRI dataset under setting 1 for the effect of $\bigp_{hf}$ and $\bigp_{nf}$.}
    \label{fig_pt-ablation}
    \vspace{-2ex}
\end{figure*}

\paragraph{More visual results.} Fig. \ref{fig_abd_ct} and Fig. \ref{fig_cmr} show comparison of qualitative results between our model and other methods on Abd-CT and CMR. Additionally, Fig. \ref{fig_loss-ablation} visualizes the effect of each auxiliary loss on our model. Furthermore, it can be seen from Fig. \ref{fig_pt-ablation} that $\bigp_{hf}$ and $\bigp_{nf}$ make different contribution to our model's performance from the perspective of segmentation center and segmentation boundary respectively.

\begin{table}[!t]
    \centering       
    \resizebox{0.48\textwidth}{!}{
    \begin{tabular}{cc|ccccc}
        \toprule
        
        \multirow{2}{*}{Setting} & \multirow{2}{*}{Pre-trained} & \multicolumn{5}{c}{Abd-MRI} \\
        & & Liver & RK & LK & Spleen & Mean \\

        \hline
        1 & ImageNet & 73.91 & 88.38 & 74.77 & 70.96 & 77.01 \\
        1 & MS-COCO & \textbf{81.55} & \textbf{90.27} & \textbf{85.58} & \textbf{79.00} & \textbf{84.10} \\
        2 & ImageNet & 71.70 & 85.98 & 71.59 & 71.01 & 75.07 \\
        2 & MS-COCO & \textbf{80.46} & \textbf{89.15} & \textbf{82.40} & \textbf{74.55} & \textbf{81.64} \\
        \bottomrule
        
    \end{tabular} 
    }
    \vspace{-2ex}
    \caption{Ablation study on Abd-MRI dataset under Setting 1 and Setting 2 for different pre-trained datasets on the performance of CoW in terms of the DICE score.}
    \label{table7}
    \vspace{-1ex}
\end{table}

\begin{table}[!t]
    \centering    
    \begin{tabular}{cc|ccccc}
        \toprule
        
        \multirow{2}{*}{$N_{hb}$} & \multirow{2}{*}{$N_{nb}$} & \multicolumn{5}{c}{Abd-MRI} \\
        & & Liver & RK & LK & Spleen & Mean \\

        \hline
        0 & 600 & 81.27 & \textbf{90.36} & 84.54 & 78.18 & 83.49 \\
        100 & 500 & 81.55 & 90.27 & 85.58 & \textbf{79.00} & \textbf{84.10} \\
        200 & 400 & 81.00 & 89.90 & \textbf{85.81} & 77.08 & 83.45 \\
        300 & 300 & \textbf{81.64} & 89.20 & 84.32 & 76.29 & 82.86 \\
        400 & 200 & 81.17 & 89.01 & 83.44 & 76.04 & 82.42 \\
        500 & 100 & 80.74 & 89.24 & 82.73 & 75.52 & 82.06 \\
        600 & 0 & 79.11 & 89.32 & 82.41 & 74.89 & 81.43 \\
        \bottomrule
        
    \end{tabular}
    \vspace{-2ex}
    \caption{Ablation study on Abd-MRI dataset under Setting 1 for the effect of number of $\bigp_{hb}$ and $\bigp_{nb}$ on the performance of CoW, where $N_{hb} + N_{nb} = 600, N_{hf} = 50$ and $N_{nf} = 50$.}
    \label{table8}
    \vspace{-1ex}
\end{table}

\begin{figure*}[!t]
    \centering
    \includegraphics[width=0.75\textwidth]{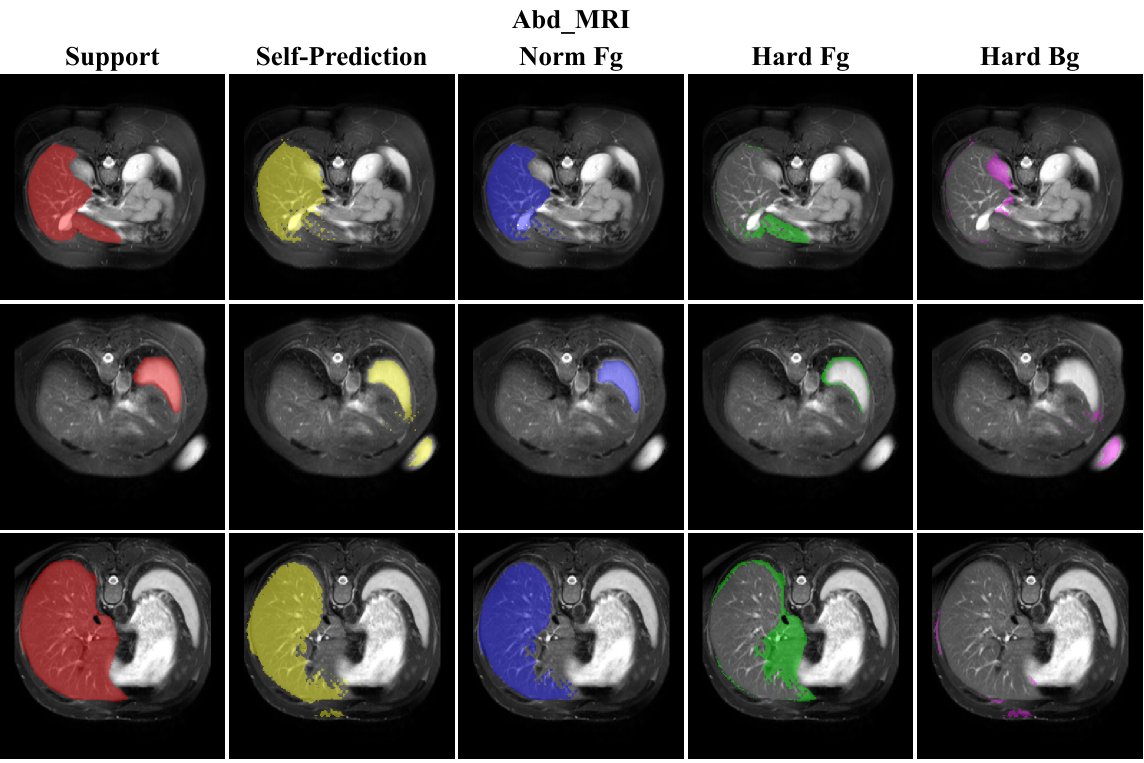}
    \caption{Visualization of features that hard and norm prototypes actually focus on during inference on Abd-MRI.}
    \label{fig_rebuttal}
    \vspace{-2ex}
\end{figure*}

\begin{table*}[!t]
    \centering  
    \resizebox{0.75\textwidth}{!}{
    \begin{tabular}{c|ccccc}
        \toprule
        
        Method & mDice & \#Params. & FLOPs & Speed & Memory-Usage\\

        \hline
        QNet & 74.75 & 18.6M & 49.8G & 40.2FPS & 12G \\
        RPT & 77.83 & 22.4M & 135.2G & 6.6FPS & 16G \\
        GMRD & 78.52 & 10.9M & 33.0G & 29.5FPS & 10G \\
        \textbf{Ours (CoW)} & 82.49 & 13.5M & 41.7G & 25.2FPS & 11G\\
        
        \bottomrule
        
    \end{tabular}}
    \caption{Efficiency comparison between our model and other methods on Abd-CT under Setting 1.}
    \label{table_rebuttal}
    \vspace{-2ex}
\end{table*}

\end{document}